\documentclass{article}

\usepackage{arxiv}

\usepackage[utf8]{inputenc} % allow utf-8 input
\usepackage[T1]{fontenc}    % use 8-bit T1 fonts
\usepackage{hyperref}       % hyperlinks
\usepackage{url}            % simple URL typesetting
\usepackage{booktabs}       % professional-quality tables
\usepackage{amsfonts}       % blackboard math symbols
\usepackage{nicefrac}       % compact symbols for 1/2, etc.
\usepackage{microtype}      % microtypography
\usepackage{lipsum}		% Can be removed after putting your text content
\usepackage{float}
\usepackage{enumitem}

\usepackage{graphicx}
\usepackage{natbib}
\usepackage{doi}

\hypersetup{hidelinks}

\title{Societal AI Research Has Become Less Interdisciplinary}

\author{ \href{https://orcid.org/0000-0001-6894-8322}{\includegraphics[scale=0.06]{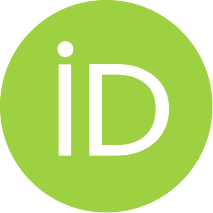}\hspace{1mm}Dror K.~Markus}%\thanks{Use footnote for providing further information about author (webpage, alternative 	address)---\emph{not} for acknowledging funding agencies.}
\\
	Department of Political Science\\
	University of Zurich\\
	\texttt{dror.markus@ipz.uzh.ch} \\
	\And
	\href{https://orcid.org/0000-0002-0635-3048}{\includegraphics[scale=0.06]{orcid.pdf}\hspace{1mm}Fabrizio  Gilardi} \\
	Department of Political Science\\
	University of Zurich\\
	\texttt{gilardi@ipz.uzh.ch} \\
    \And
	\href{https://orcid.org/0000-0002-3698-4414}{\includegraphics[scale=0.06]{orcid.pdf}\hspace{1mm}Daria Stetsenko} \\
	Department of Computational Linguistics\\
	University of Zurich\\
	\texttt{daria.stetsenko@uzh.ch} \\
}

\renewcommand{\headeright}{}
\renewcommand{\undertitle}{}
\renewcommand{\shorttitle}{Societal AI Research Has Become Less Interdisciplinary}

\hypersetup{
pdftitle={A template for the arxiv style},
pdfsubject={AI, Artificial Intelligence},
pdfauthor={Dror K. Markus, Fabrizio Gilardi, Daria Stetsenko},
pdfkeywords={AI Research, Responsible AI, AI Governance, Interdisciplinarity, Ethics, 
Computational Social Science}
}

\begin{document}
\maketitle
\renewcommand{\thefootnote}{\fnsymbol{footnote}}
\footnotetext[1]{All data and replication materials will be made publicly available upon publication. Until then, they are available from the authors upon request.}
\renewcommand{\thefootnote}{\arabic{footnote}}
\begin{abstract}
   As artificial intelligence (AI) systems become deeply embedded in everyday life, calls to align AI development with ethical and societal values have intensified. Interdisciplinary collaboration is often championed as a key pathway for fostering such engagement. Yet it remains unclear whether interdisciplinary research teams are actually leading this shift in practice. This study analyzes over 100,000 AI-related papers published on ArXiv between 2014 and 2024 to examine how ethical values and societal concerns are integrated into technical AI research. We develop a classifier to identify societal content and measure the extent to which research papers express these considerations. We find a striking shift: while interdisciplinary teams remain more likely to produce societally-oriented research, computer science–only teams now account for a growing share of the field’s overall societal output. These teams are increasingly integrating societal concerns into their papers and tackling a wide range of domains - from fairness and safety to healthcare and misinformation.These findings challenge common assumptions about the drivers of societal AI and raise important questions. First, what are the implications for emerging understandings of AI safety and governance if most societally-oriented research is being undertaken by exclusively technical teams? Second, for scholars in the social sciences and humanities: in a technical field increasingly responsive to societal demands, what distinctive perspectives can we still offer to help shape the future of AI?

\end{abstract}

% % keywords can be removed
% \keywords{AI Research \and Responsible AI \and AI Governance \and Interdisciplinarity \and Ethics \and Computational Social Science}

\section{Introduction}
Recent years have witnessed rapid advances in artificial intelligence (AI), especially with the rise of generative models. These developments have led to a growing awareness of the technology's transformative potential across sectors and domains \citep{harari2017reboot}. This includes promising opportunities, but also sparks growing concern over ethical issues and risks, such as algorithmic bias, automation, sustainability and misinformation \citep{Weidinger:2023a, hanna2023aiharms, gilardi2024narratives, Kasirzadeh:2025a}. This societal concern has permeated the AI research community as well \citep{aiethics2022}. Prominent AI scholars have taken to public platforms to highlight ethical challenges such as bias in large language models and facial recognition technologies \citep{boyle2020uw}. These high-profile discussions reflect broader undercurrents in the AI research community. Surveys consistently reveal a shared belief in the importance of ethical AI development and concern with corresponding risks \citep[e.g.,][]{grace2024thousands,michael2023nlp,zhang2021ethics}. 

These growing concerns increasingly direct the structure and direction of AI research itself, manifesting in at least three major strands of scholarly activity. One centers on the development of \textit{responsible AI}, an umbrella term that encompasses design that prioritizes harm prevention and societal welfare \citep{Voeneky_Kellmeyer_Mueller_Burgard_2022}. Research in this area focuses on principles such as transparency, explainability, fairness, accountability, human oversight, and privacy. Scholars working in this space often aim to identify violations of these principles and propose technical or procedural solutions to mitigate potential harms \cite[e.g.,][]{weidinger2023veil}. A second strand involves the application of AI to pressing public problems \citep[e.g.,][]{ashurst2022disentangling}. Here, researchers seek to harness AI technologies to address issues such as misinformation, hate speech, climate change, or public health \citep[e.g.,][]{kotarcic-etal-2022-human, Li2023, ron-etal-2023-factoring}. Finally, a third stream of scholarship centers on \textit{AI governance}, examining how institutions, policies, and funding structures can shape the development and deployment of AI in ways that align with democratic values and the public interest \cite[e.g.,][]{Burri_2022, zaidan2024aigovernance}. Together, these channels demonstrate the expanding scope of societal orientation in AI research. 

The AI research community has been implementing internal institutional responses to foster more societally-oriented projects. Leading AI conferences, such as the Conference on Neural Information Processing Systems (NeurIPS) and the Association for Computational Linguistics (ACL), have begun asking authors to include ethical or societal impact statements in their research submissions \citep{ashurst2022ai,benotti2022ethics,karamolegkou2024ethical}. Additionally, smaller, targeted workshops focusing on issues such as misinformation, hate speech detection, and fairness are becoming fixtures at major conferences. In another step, the establishment of the Ethics and Society Review (ESR), is attempting to condition project funding on researchers’ engagement with ethical and societal considerations \citep{bernstein2021esr}.

However, recent studies have raised concerns about the depth and efficacy of these mechanisms. For example, researchers often opt out of submitting required ethics or broader impact statements, or they provide only minimal content, treating the task as a procedural formality rather than an opportunity for meaningful reflection \citep{ashurst2022ai, abuhamad2020likeresearcherstatingbroader, do2023}. Even when statements are submitted, they frequently emphasize potential benefits while avoiding critical discussion of risks or harms \citep{birhane2021values, nanayakkara2021}. This tendency toward optimistic framing may reflect strategic considerations, as researchers anticipate how such content might be evaluated during peer review \citep{prunkl2021}. At the same time, these patterns may point to deeper structural challenges for societal orientation: researchers working on theoretical or foundational projects often struggle to anticipate downstream consequences, particularly when those consequences are diffuse or indirect \citep{nanayakkara2021, prunkl2021}. Compounding these issues, many computer scientists report uncertainty about how to engage meaningfully with ethical concerns due to limited training or a lack of field-specific guidance \citep{do2023}. Together, these findings suggest that while institutional mechanisms have raised awareness, there is still a long way to go in fostering the substantive integration of societal considerations into AI research practice.

Our study builds on this literature, but shifts attention to a different mechanism: interdisciplinary collaboration. If computer scientists might describe themselves as having difficulty in engaging with or conceiving of downstream implications, the inclusion of diverse disciplinary perspectives could offer a solution. We see repeated emphasis in policy papers and institutional guidelines for more interdisciplinary research teams as a key strategy for aligning AI with societal values \cite[e.g.,][]{montag2024ai, zhang2021ethics, oecd2019principles}. By including researchers from diverse fields - such as political science, philosophy, and law - it is assumed that the research will be more likely to integrate ethical, social, and political dimensions into projects \cite{hecht2020, montag2024ai}. The rationale for this stems from two factors. First, interdisciplinary teams are often assumed to better reflect real-world problem-solving by drawing on diverse forms of expertise and experience \cite{hu2024interdisciplinary}. These teams are seen as particularly well-suited to addressing AI's multifaceted societal impacts, focusing research on concrete issues or questions. Second, interdisciplinary teams are more prone to disrupting entrenched research paradigms \cite{chen2024interdisciplinarity, krauss2024science}. In AI fields traditionally dominated by performance benchmarks, disciplines outside of computer science might play a crucial role in introducing societal considerations and targets. 

In this paper, we examine the prevalence of \textit{societal orientation} in technical AI research. We define societal orientation as the integration of ethical values and societal concerns that shape the motivations, design and stated outcomes of research (more in Section~\ref{sec:societalinfluence}). This concept reflects how researchers orient towards studies with broader social questions. We then examine the role of one widely promoted mechanism-interdisciplinary collaboration-in advancing such societally-oriented research. To explore these dynamics, we utilize a unique dataset of over 100,000 technical AI research papers, complemented by author metadata used to infer each team's disciplinary composition. Additionally, we develop a supervised machine learning classifier to detect the societal orientation of each paper, and a pipeline to identify and classify the main research question of each paper.

We extend previous research in several key ways. First, we apply a large-scale computational analysis to move beyond specific research subfields \cite[e.g.,][]{birhane2021values,scheuerman2021datasets}, and instead examine multiple AI domains together. Additionally, as opposed to examining only impact statements \cite[e.g.,][]{ashurst2022disentangling, benotti2022ethics, karamolegkou2024ethical}, by examining full papers, we enable a more fine-grained assessment of their societal orientation. Crucially, we also evaluate the efficacy of interdisciplinarity, a specific mechanism generally considered to be a driver of societal research. 

Our analysis reveals a nuanced picture. Interdisciplinary teams are indeed more likely to produce societally-oriented work. However, despite this, research focusing on such concerns and issues is being dominated by computer science–only teams. These teams increasingly center societal issues in their research questions and contribute across diverse relevant topics. We cite three potential explanations for this shift: evolving norms within the computer science community, a broader transition from foundational to applied research, and the rise of computational social science as a hybrid subfield. As public discourse on AI's societal implications intensifies, this study offers a timely empirical assessment of mechanisms assumed to help align technological development with societal concerns and needs. More broadly, this study contributes to ongoing debates about how best to foster societally-aligned science in fast-moving technological domains. 

\section{Research Design}

\subsection{Data}
We constructed a dataset of 103,314 research papers from the ArXiv repository using the public API \citep{clement2019usearxivdataset}. ArXiv includes both preprints and peer-reviewed conference or journal publications, offering a broad snapshot of ongoing AI research. We focus on ArXiv because it captures work produced by the technical research community most actively developing AI systems—distinct from scholarship on societal applications or implications published in field-specific journals.

We selected papers from four ArXiv subfields: Artificial Intelligence, Computer Vision, Computational Linguistics, and Computers and Society.\footnote{Labels are assigned by authors upon submission.} These domains lie at the intersection of technical innovation and societal relevance, where issues such as ethics, governance, and public impact are more likely to be addressed. We focused on papers published between 2014 and 2024—a period that coincides with major advances in deep learning and the growing prominence of AI-related societal debates. After removing duplicate versions of the same paper, the final dataset comprised 101,919 unique research articles. Further details on downloads, preprocessing, and filtering are provided in Section \ref{Matmethods-Data}.

To assess the disciplinary composition of each research team, we used the Semantic Scholar API to retrieve the publication history of every author listed on each paper in our corpus \citep{Kinney2023TheSS}. Semantic Scholar assigns each publication a ‘field of study’ corresponding to its academic discipline. We extracted the field of study for each one of a researcher's previous research papers, and mapped it to one of three umbrella categories: \textit{Computer Science and Engineering} (CS), \textit{Natural Sciences and Medicine} (NSM), or \textit{Social Sciences and Humanities} (SSH). For each author, we then tallied the number of publications falling under each umbrella category, producing a distribution of disciplinary activity over their prior work.

Based on this distribution, we assigned each author a final disciplinary label. If at least 90\% of an author's prior publications belonged to a single umbrella category, they were assigned that category.\footnote{We implemented a robustness check on this threshold, detailed in Appendix A} Otherwise, the author received a dual label reflecting their two most frequent categories. 

Papers were then classified into four team types: (1) \textit{CS-only}, if all authors were labeled exclusively as computer scientists or engineers; (2) \textit{Interdisciplinary (SSH)}, if the team included at least one author with an SSH label and none with an NSM label; (3) \textit{Interdisciplinary (NSM)}, if the team included at least one NSM-labeled author and none with SSH; and (4) \textit{Fully Interdisciplinary}, if authors from all three umbrella categories were represented. This classification enables us to systematically examine the relationship between team composition and the manifestation of societal considerations in AI research.

\subsection{Two Measures of Societal Orientation} \label{sec:societalinfluence}
A central component of this study is the measurement of the \textit{societal orientation} of AI research. We define this as the integration of ethical values and societal concerns into a paper’s research motivations, design, or stated outcomes. This encompasses two components: normative principles—such as fairness, accountability, safety, and privacy—as well as substantive engagement with topics like healthcare, misinformation, or environmental sustainability. Together, these perspectives cover the three strands of societal research as noted above. To operationalize this construct, we developed a codebook grounded in prior work identifying ethical values in AI research \citep{ashurst2022ai}, as well as survey results of researcher concerns \citep{grace2024thousands, zhang2021ethics}. Annotators were trained using this codebook to identify sentence-level indicators of societal orientation across our diverse set of AI subfields.

To enable large-scale analysis, we trained a logistic regression classifier on 1,000 manually annotated sentences. For each paper, we extracted only the \textit{Abstract}, \textit{Introduction} and \textit{Conclusion} sections as we assumed these sections were most relevant for societal motivations or discussion, as opposed to other, more technical sections. Annotation was performed by two independent coders, with a third researcher resolving disagreements. Sentences were embedded using SciBERT, a transformer-based language model fine-tuned for scientific text \citep{Beltagy2019SciBERT}. After evaluating multiple model architectures and hyperparameters, the final model achieved an F1 score of 0.93. Additional details on model development, validation, and robustness are provided in the Methods section. 

The classifier detects whether a given sentence expresses societal orientation. For example, one sentence reads: “While a common goal of AI is to work towards more human-like (anthropomorphic) agents, a challenge that remains is to consider the trade-off between the naturalness of a system and the safety of its deployment.” This reflects a concern with the ethics of AI safety - issues such as transparency of models, reliability of outputs and possible misuse of such capabilities. Another sentence illustrates orientation towards societal topics and applications: “Several attempts have been made to classify hate speech using machine learning but the state-of-the-art models are not robust enough for practical applications.” Here, the text emphasizes the research's link to real-world issues or applications. These sentence-level classifications were aggregated for each document to generate the first metric of societal orientation - the percentage of a paper's content (specifically, of its \textit{Abstract}, \textit{Introduction} and \textit{Conclusion} sections) expressing such considerations.

Beyond this paper-level proportion of societal orientation, we developed a second metric: whether the central focus of the paper was of societal dimensions. The rationale behind this variable was to identify if the societal orientation was central to the research project, or introduced only peripherally, for example, through task selection, dataset choice, or a brief concluding remark. To measure this, we extracted the main research question from each paper using a large language model, prompted with the title and abstract. The resulting questions were then classified using our societal orientation model to determine whether they reflected a societal concern (details in Materials and Methods Section). Together, these two variables provide fine-grained measures of societal orientation in research - allowing us to explore research whose central focus are societal considerations or topics, as well as the diverse ways in which ethical or societal concerns are integrated into AI research motivations and design even peripherally. 

\section{Results}

%% Add descriptive stats
Having run our classifier on the documents in our corpus, we find several noteworthy descriptive insights. First, societally-focused papers are relatively rare, but their prevalence is meaningful considering the technical nature of our corpus. When analyzing the main research questions for each paper, we find that 12\% of our corpus's questions were classified as being societal-oriented. Such limited emphasis is clear at the sentence-level as well. Only 26\% of all sentences (from the \textit{Abstract}, \textit{Introduction} and \textit{Conclusion} sections) were classified as societal, and these sentences mostly co-occur within a small proportion of research papers. Additionally, the percent of societal-orientation has fluctuated over our timeframe, without clear temporal trends.

\subsection{Interdisciplinary Teams and Societal Orientation}

We examined whether interdisciplinary collaboration is associated with greater integration of societal considerations in AI research. As described in the methods, we classify each paper (\textit{Abstract}, \textit{Introduction} and \textit{Conclusion} sections) into one of four team types based on the disciplinary backgrounds of its authors: CS-only teams, interdisciplinary teams including researchers from the natural sciences or medicine (NSM), interdisciplinary teams including social scientists or humanists (SSH), and fully interdisciplinary teams spanning all three domains.

\begin{figure}%[htbp]
    \centering
    \includegraphics[width=0.8\linewidth]{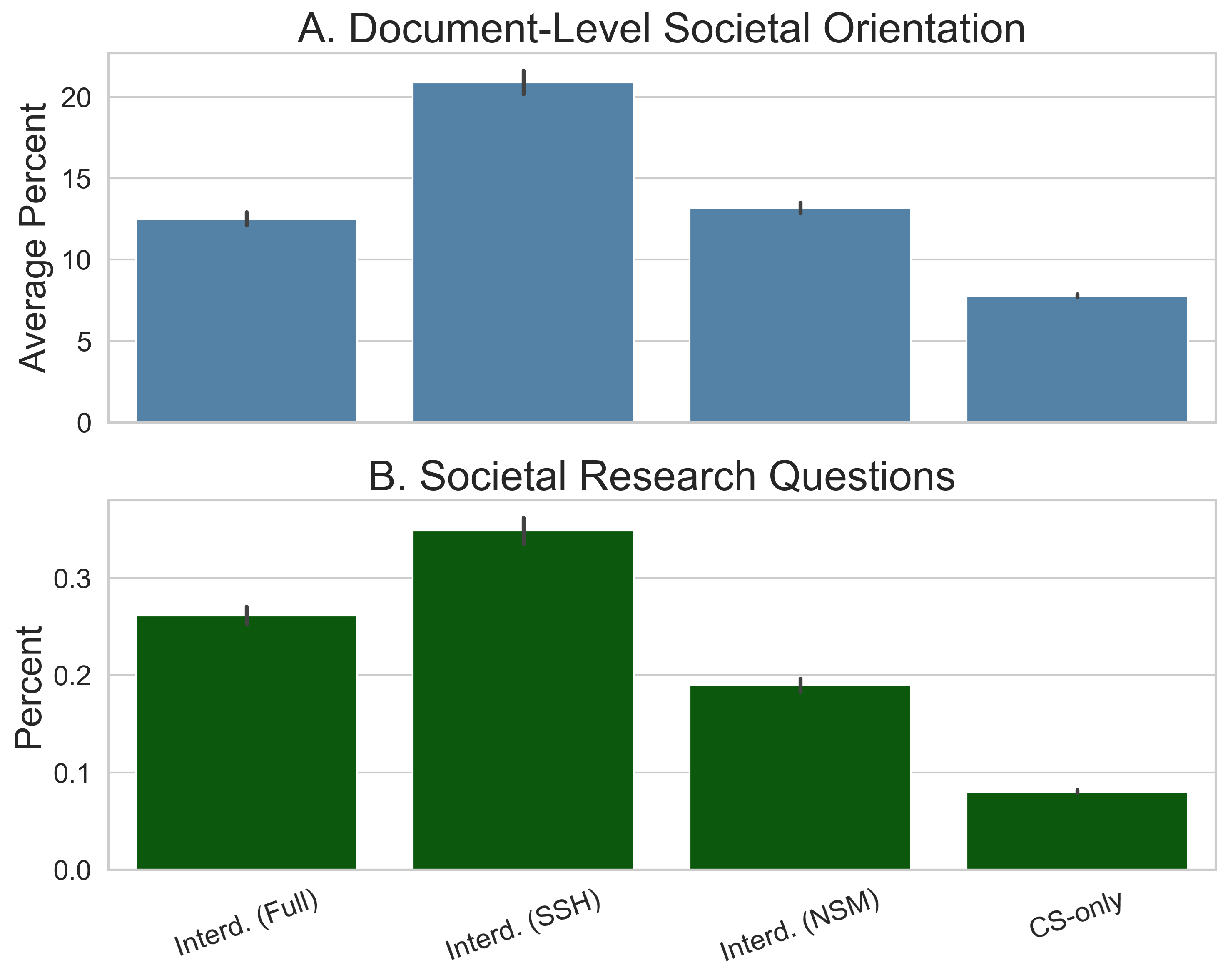}
    \caption{\textbf{A.} Average proportion of societally-oriented sentences per paper (the \textit{Abstract}, \textit{Introduction} and \textit{Conclusion} sections), by team type. \textbf{B.} Share of papers whose research question centers on a societal issue, by team type. Error bars represent 95\% confidence intervals.}
    \label{fig:avg_societal_engagement_teamtype}
\end{figure}

Figure~\ref{fig:avg_societal_engagement_teamtype} presents the average levels of societal orientation for each team type. Panel A shows results based on our sentence-level classifier, which captures the proportion of each paper that explicitly expresses ethical or societal concerns. Interdisciplinary teams score markedly higher on this measure compared to CS-only teams. Papers authored by teams including social scientists or humanists (SSH) exhibit the highest levels of societal orientation, with an average of 20.9\% of sentences flagged as societally-oriented. This is nearly three times the average for CS-only teams (7.8\%). Fully interdisciplinary teams and NSM-inclusive teams also show elevated levels of societal orientation (12.5\% and 13.2\%, respectively). Panel B shows a similar pattern based on our second measure: whether a paper’s main research question reflects a societal issue. Here too, interdisciplinary teams lead. Papers by SSH-inclusive teams are most likely to focus on societal concerns (35\%). By contrast, only 8\% of CS-only papers center a societal research question.

These descriptive patterns suggest that teams that include contributors from the social sciences, humanities, or natural sciences are significantly more likely to integrate societal values or tackle societal questions. These patterns lend empirical support to prevailing policy recommendations that promote interdisciplinary teams as a mechanism to foster societally-oriented AI research. Such results remain robust even when using different thresholds for author affiliation and classifier sensitiviy (detailed in Appendix A). A fixed-effects panel regression model also confirms that these differences remain robust when controlling for year, subfield, team size, and article length (Appendix B).

\begin{figure}%[htbp]
    \centering
    \includegraphics[width=1\linewidth]{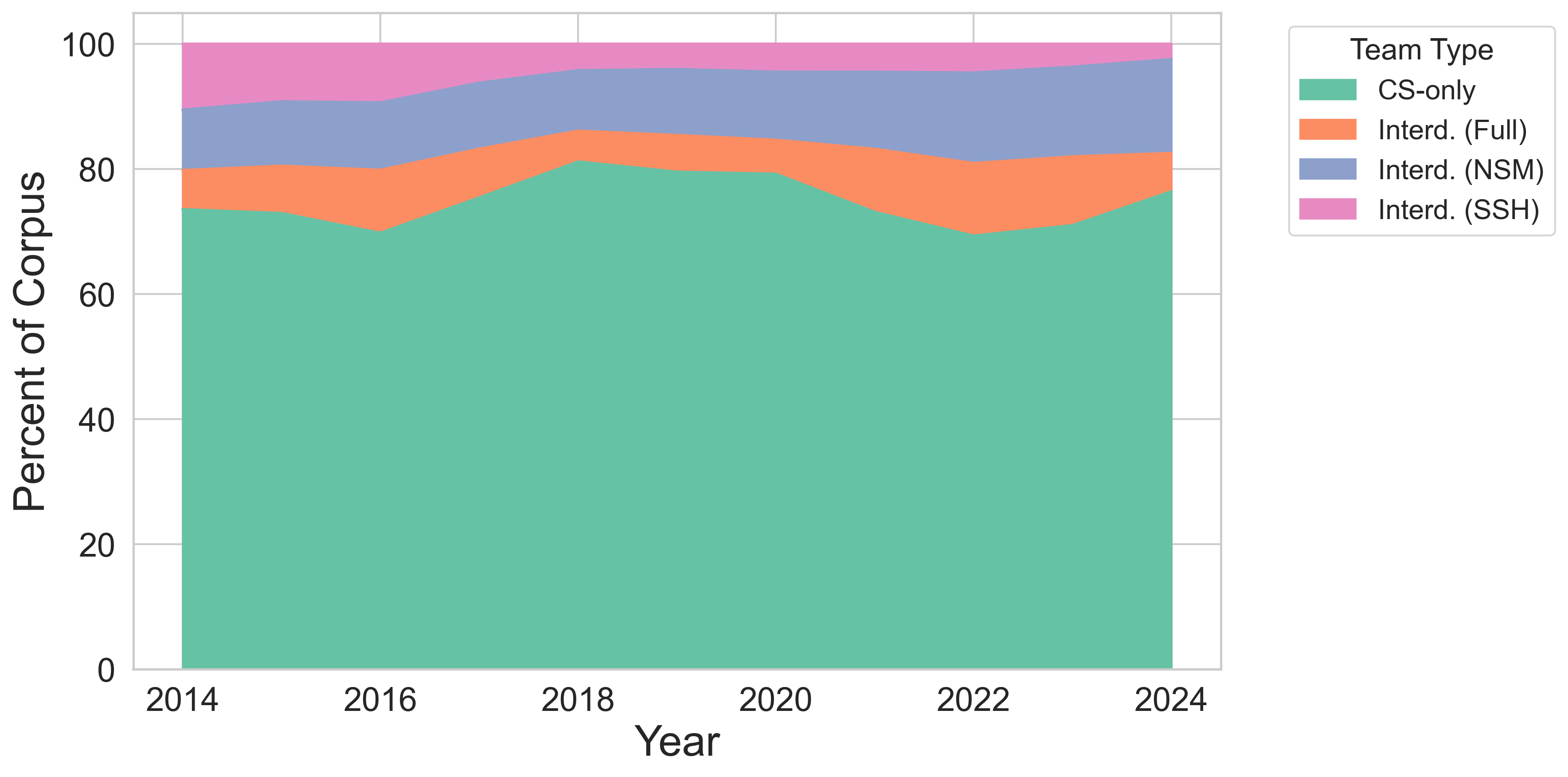}
    \caption{Proportion of team types in AI research, 2014–2024. Each line represents the annual share of papers produced by a given team type, calculated as a proportion of the total number of papers published that year. The results show that CS-only teams consistently comprise the majority of research output across all years and remain stable over time.}
    \label{fig:team_composition_over_time}
\end{figure}

However, this pattern tells only part of the story. As shown in Figure~\ref{fig:team_composition_over_time}, when we examine team composition over time, we find that the share of interdisciplinary teams has remained relatively stable at about 75\%. A linear trend analysis reveals no statistically significant change in the share of CS-only teams over the past decade, confirming this stagnation ($\beta = -0.013$, $p = 0.975$). 

\begin{figure}%[htbp]
    \centering
    \includegraphics[width=1\linewidth]{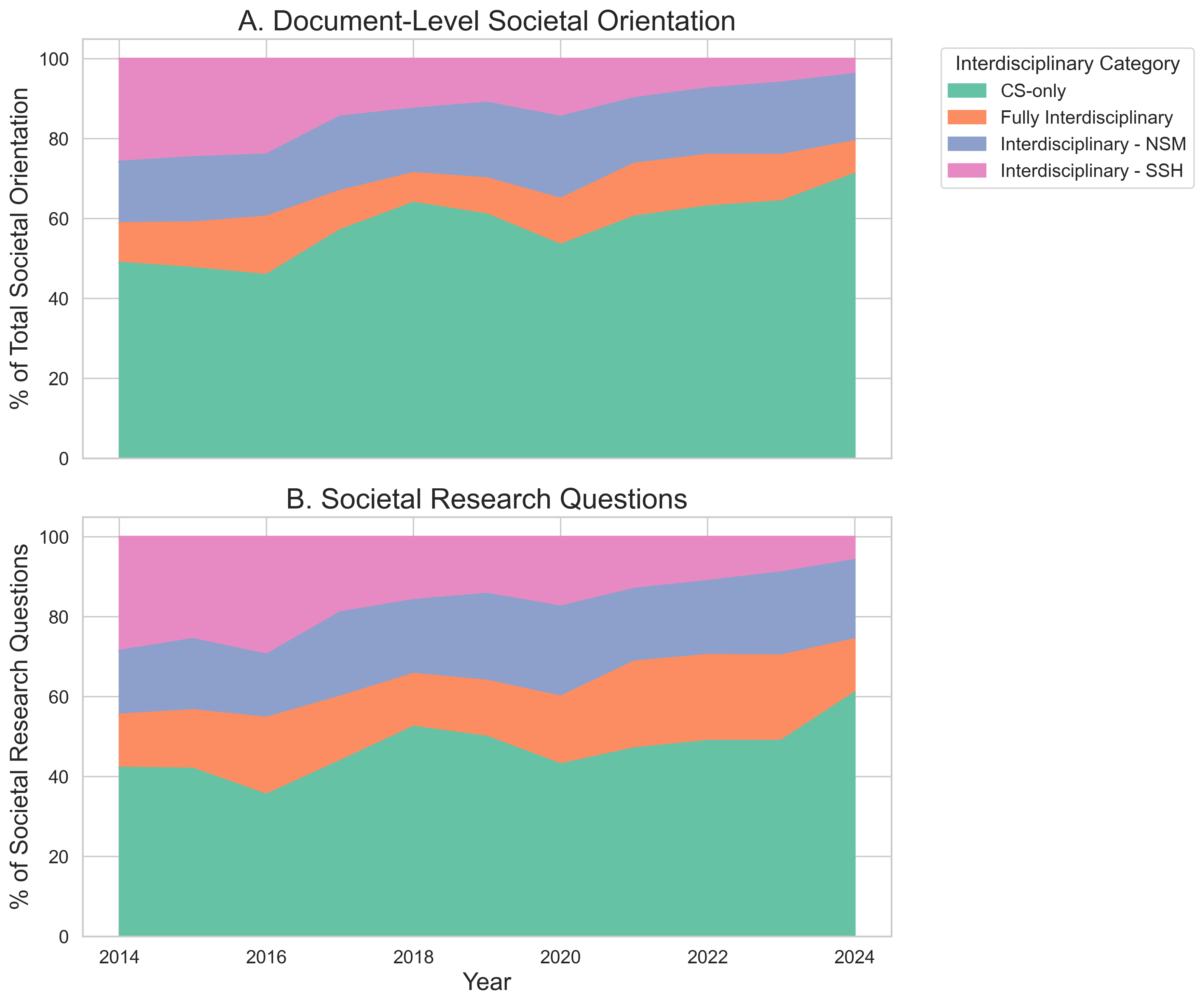}
    \caption{
    \textbf{A.} Share of total sentence-level societal orientation attributable to each team type, by year. \textbf{B.} Share of papers with societal research questions attributable to each team type, by year. Both panels use stacked area charts, with the y-axis representing the proportion of all societally-oriented content (Panel A) or papers (Panel B) in each year. Both results show a clear and consistent rise in the contribution of CS-only teams over time. 
    }
    
    \label{fig:cs_societal_contribution}
\end{figure}

However, we do find a trend whereby CS-only teams are increasingly focusing on societal questions and integrating societal considerations. While they remain less societally-oriented on average, CS-only teams are now responsible for a growing share of the field’s overall societal output. As shown in Figure~\ref{fig:cs_societal_contribution}, the proportion of all sentence-level societal orientation attributable to CS-only teams rose from 49.0\% in 2014 to 71.2\% in 2024. A linear regression confirms this upward trend is statistically significant ($\beta = 2.05$, $p = 0.001$). Panel B reveals a similar pattern for our second measure: the share of societally-framed research questions also increasingly originates from CS-only teams. Together, these results underscore a structural shift in the locus of societal orientation, suggesting that evolving norms and practices within computer science itself are playing a leading role in driving this growth.

Such changes become especially clear when focusing on teams that include researchers from the social sciences and humanities (SSH)—the group most strongly associated with societally-oriented work. As shown in Figure~\ref{fig:team_composition_over_time}, the relative share of papers produced by SSH-inclusive teams has declined over time. A parallel pattern appears in Figure~\ref{fig:cs_societal_contribution}, where their contribution to the field’s societal orientation has dropped sharply. In 2014, SSH-inclusive teams accounted for 25.7\% of all societally-oriented sentences; by 2024, this figure had fallen to just 3.8\%. A similar decline is evident in societally focused research questions, where SSH teams’ share decreased from 28.5\% to 5.8\% over the same period. These patterns reflect the broader reconfiguration in the field’s societal orientation of research. 

\subsection{CS-only Teams Tackling Full Range of Societal Topics}

The dominance of CS-only teams led us to wonder if there were important, distinctive contributions of interdisciplinary teams. In particular, we asked whether interdisciplinary teams are more likely to incorporate certain themes—such as misinformation, public health, or climate change—than their CS-only counterparts. We began by extracting all sentences labeled as societally-oriented by our classifier and applied topic modeling to group them into semantically coherent themes (see Methods). Each paper was then assigned a set of topic relevance scores based on the average similarity between its societally-oriented content and each topic.

\begin{figure}%[htbp]
    \centering
    \includegraphics[width=0.9\linewidth]{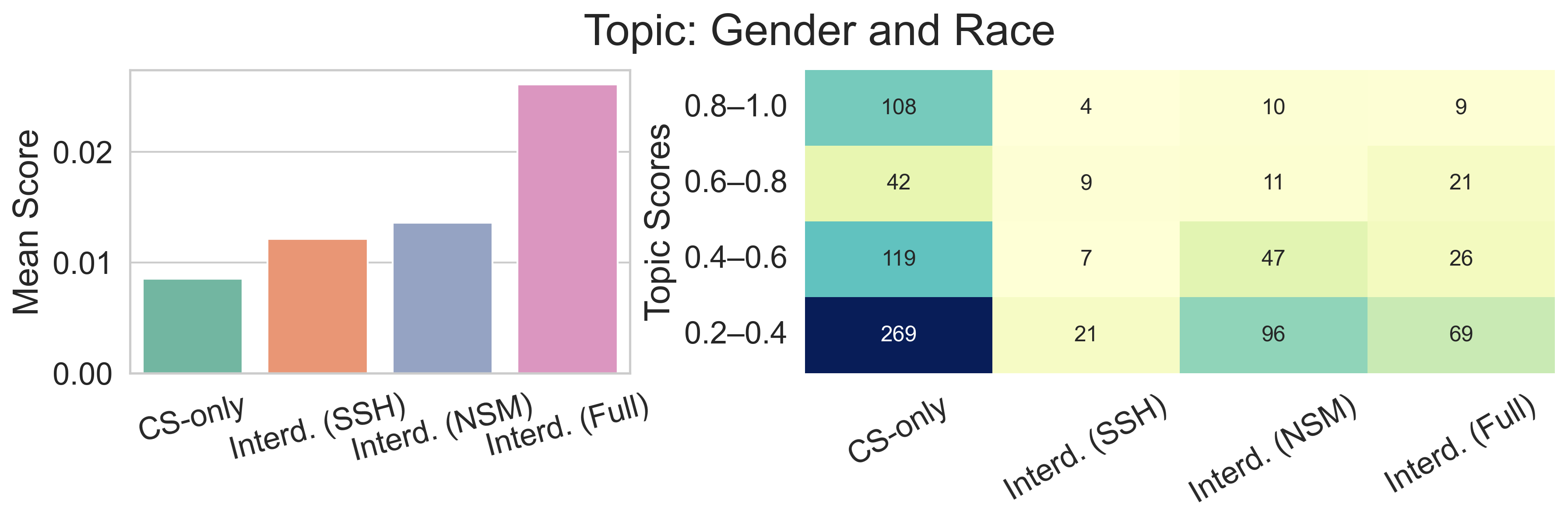}
    \includegraphics[width=0.9\linewidth]{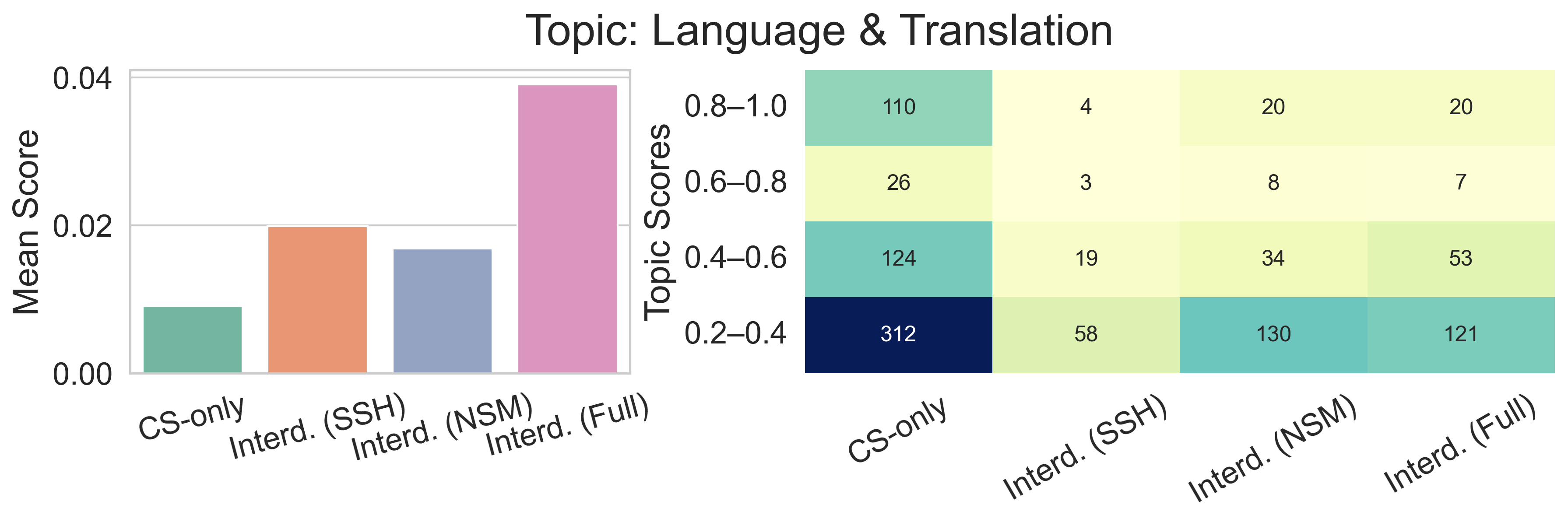}
    \includegraphics[width=0.9\linewidth]{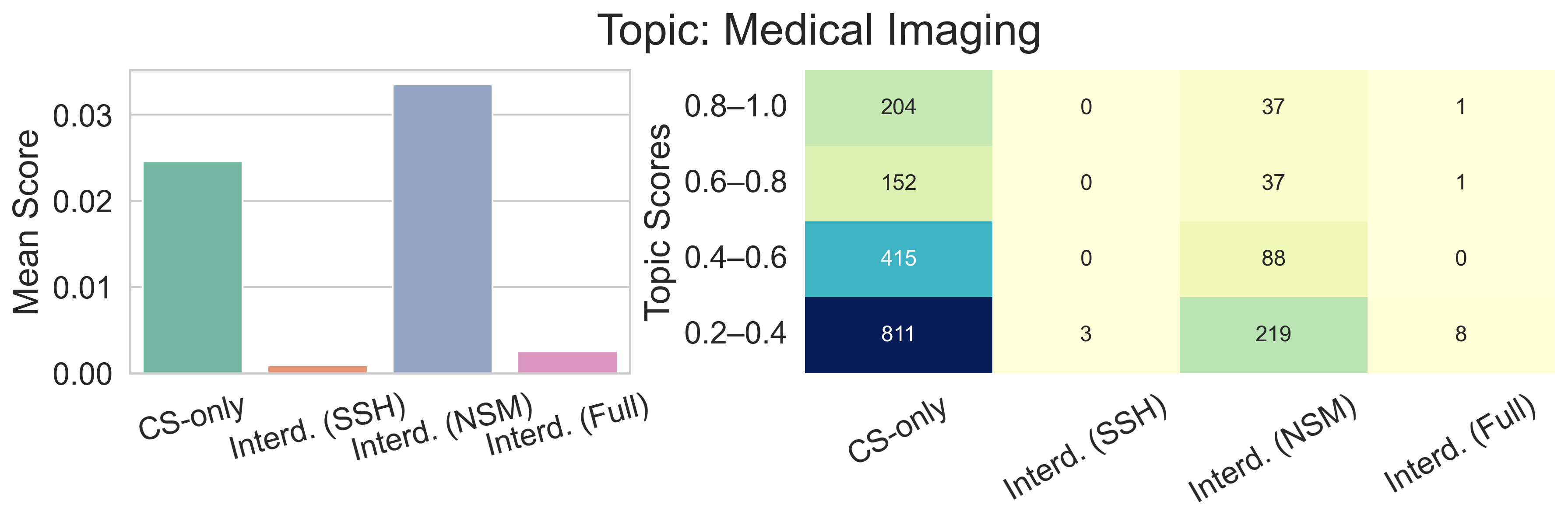}
    \caption{Topic-specific societal orientation by team type. Each row represents one of the four most frequent societal topics in the dataset. \textbf{Panel A (left)} shows the mean topic relevance score for each team type, indicating average orientation levels. \textbf{Panel B (right)} displays a heatmap of document counts, with each cell showing the number of papers in a given team category that fall within four topic relevance score bins (0.2–0.4, 0.4–0.6, 0.6–0.8, and 0.8–1.0). This layout enables comparison of both mean orientation and the distributional breadth of contributions across team types. While interdisciplinary teams - especially those involving social sciences or natural sciences - often achieve the highest average topic relevance, CS-only teams consistently contribute the largest volume of highly relevant papers. For instance, in \textit{Gender and Race}, even though the Fully Interdisciplinary teams have the highest mean association with the topic, CS-only teams account for the majority of papers with strong topical alignment, underscoring their growing role in addressing a wide range of societal issues.}
    \label{fig:topic_examples}
\end{figure}

To compare contributions across team types, we selected three of the most frequent topics in the dataset - \textit{Gender and Race}, \textit{Language and Translation}, and \textit{Medical Imaging} - and visualized them in Figure~\ref{fig:topic_examples}. Each topic is presented using two panels: the left panel shows the mean topic relevance score per team type, while the right panel displays a heatmap of the number of papers falling into different score ranges (thresholded at 0.2 and above). As expected, interdisciplinary teams - particularly those including social scientists or natural scientists - tend to achieve the highest average topic relevance scores. However, when we examine the heatmaps, a consistent pattern emerges: CS-only teams produce the most high-scoring papers across all topics. Even in domains where interdisciplinary expertise might seem most essential, CS-only teams are not only participating - they often contribute a large share of the top-relevance papers. This reinforces a recurring finding throughout our analysis: CS-only teams are increasingly leading the field’s orientation towards societal concerns, both in volume and in topical breadth.

\section{Conclusion}
Together, these findings provide a window into the integration of societal concerns into AI research, and the role interdisciplinary teams play in such work. Mixed disciplinary teams, as widely assumed, are more likely to focus on societal dimensions of AI in their research. However, CS-only teams are now tackling such questions on their own, dominating societal-oriented work across a variety of topics. This is particularly pronounced when considering that the proportion of interdisciplinary teams has remained stable, pointing to an internal shift occurring in the computer science community. 

We propose several possible explanations for this trend, which could be explored further in future work. First, the growth in societal orientation may reflect the success of institutional interventions—such as the introduction of social impact statements, the growth of societally focused workshops, and the conditioning of funding on ethical considerations. This interpretation offers a more optimistic perspective than prior studies, which have questioned the depth and efficacy of such mechanisms.

A second possibility is that the observed increase in societal orientation reflects a natural evolution of the field itself—from an emphasis on foundational methods toward real-world applications. Having seen several dramatic breakthroughs in AI technologies, researchers may turn to explore their implications across diverse domains and addressing varied questions. This is especially true with the rise of LLMs, whose general-purpose capabilities have opened new pathways for applied research \cite{movva-etal-2024-topics}. As these models reduce the need for domain-specific architectures or handcrafted features, researchers may increasingly shift focus from foundational innovation toward addressing real-world challenges across sectors.

Finally, the results may signal the emergence of computational social science as a nascent but growing interdisciplinary space. Although this community is not yet clearly delineated or institutionally grounded, an increasing body of work focuses on the synthesis of computational tools with social science questions. Researchers working in this space might still categorized as computer scientists, due to their educational or professional trajectories, even when their research engages deeply with societal concerns.

Beyond these points for further exploration, there are two key limitations in the current study that could be addressed in future work. First, our measure of societal orientation is relatively broad and uni-dimensional. While it successfully captures general expressions of ethical and societal concern, it does not differentiate between distinct types of values or assess the stance taken toward them. Future work could build on this by incorporating sentiment analysis or more nuanced classification schemes to distinguish between issue types and researcher positions. Such fine-grained analysis would enable clearer comparisons between the concerns emphasized by AI researchers and those raised in public or policy discourse. Second, our analysis focuses on ArXiv, a public preprint repository that—while rich and representative of academic AI research—excludes much of the work conducted in corporate, government, or military settings. These settings often lack transparency but may nonetheless drive the development and deployment of impactful AI systems. Future research should explore alternative sources such as industry white papers or press releases to expand coverage.

This paper offers a first large-scale, quantitative examination of one widely cited mechanism for promoting societal alignment in AI research: interdisciplinary collaboration. Drawing on a comprehensive corpus of technical AI papers spanning multiple subfields, we analyze the research being developed at its core - where AI is being developed, not merely applied. We confirm that interdisciplinary teams are indeed more likely to integrate societal considerations into their work. At the same time, we document an important internal shift: computer science–only teams are increasingly addressing these concerns on their own and now constitute the majority of societally-oriented AI research.

These findings raise several important questions. On the one hand, they may reflect a positive development - namely, the emergence of new norms within computer science that prioritize ethical and societal dimensions. On the other hand, they prompt concerns about what might be lost in the absence of diverse disciplinary perspectives. Are important voices, frameworks, or critical perspectives being sidelined? At the same time, they pose an important challenge for scholars from other disciplines—to articulate and deliver distinct contributions to the ongoing development of AI. As computational social scientists ourselves, these results prompt self-reflection. It is not enough to merely raise social or ethical questions; we must strive to offer distinctive contributions that complement and enhance ongoing technical work. 

One avenue for such contribution may lie in bridging the gap between research-internal conceptions of societal relevance and the concerns of broader publics. While our study assesses the societal orientation of research, it does not address how these concerns align with actual public demands or democratic interests. Prior work suggests that AI researchers and members of the public often diverge in their prioritization of risks, benefits, and values \citep{karamolegkou2024ethical}. This may represent an opportunity - and a responsibility - for social scientists to better align academic inquiry with democratic representation. This point is supported by a recent survey finding AI researchers seeking "human values" for the technology, while still paying limited attention to insights coming from the social sciences \cite{odonovan2025visions}. Rectifying this disconnect may be essential to ensuring that the future of AI research is not just societally-aware, but democratically responsive.

\section{Methods}
\subsection*{Data Collection and Preprocessing}
\label{Matmethods-Data}
We collected 101,919 AI-related papers from the ArXiv repository. We selected papers from 2014–2024 across four categories - \textit{Cs.Cl}, \textit{Cs.Cv}, \textit{CS.Cy}, and \textit{Cs.Ai}. The list of relevant research papers was retrieved first from the Arxiv API. We then downloaded the pdf files from the Kaggle repository at \url{https://www.kaggle.com/datasets/Cornell-University/arxiv}.  We then extracted the Abstract, Introduction and Conclusion sections from each research paper's pdf file. We note that for some papers, the one or more of these sections might be missing due to issues with the pdf file or text formatting. 86\% of the papers were missing abstracts, 18\% were missing their introduction section, and 35\% were missing the conclusion. While these numbers seem high, in reality only 35 research papers were missing all three sections. Thus, for each paper we still had at least one large section of text with which to analyze the societal orientation. Additionally, in our analysis, we normalize all article lengths by measuring the percent of a paper's text containing societal orientation.

\subsection*{Disciplinary Inference}
We used the Semantic Scholar API to retrieve the publication history of every author listed on each paper in our corpus \citep{Kinney2023TheSS}. Semantic Scholar assigns each publication a ‘field of study’ corresponding to its academic discipline. We extracted the field of study for each one of a researcher's previous research papers, and mapped it to one of three umbrella categories: \textit{Computer Science and Engineering} (CS), \textit{Natural Sciences and Medicine} (NSM), or \textit{Social Sciences and Humanities} (SSH). For each author, we then tallied the number of publications falling under each umbrella category, producing a distribution of disciplinary activity over their prior work. Based on this distribution, we assigned each author a final disciplinary label. 

\subsection*{Annotation and Classifier Training}
To train a classifier for societal orientation, we sampled sentences from the full corpus of over 14 million AI-related sentences. We first filtered out sentences with fewer than five words, which are unlikely to contain substantive content. We then implemented a two-pronged sampling strategy for annotation. First, we randomly sampled $n=167$ sentences from each of three major AI subfields (Natural Language Processing, Computer Vision and AI, and AI and Society). This step ensured generalizability. Second, we used a guided sampling approach to enrich the dataset with societally relevant content. Specifically, we filtered for sentences containing one or more expert-curated keywords associated with ethical or societal values, such as “fairness,” “privacy,” “sustainability,” and “misinformation.” We randomly sampled up to $n=167$ of these keyword-matched sentences per subfield. The combined dataset was shuffled and deduplicated to produce a final annotation set of 1,002 unique sentences, equally distributed across subfields and sampling strategies.

We then asked two research assistants to annotate the sentences based on a detailed codebook grounded in prior work (detailed in Appendix C), with one of the paper authors acting as a tie-breaker in such cases. These formed the basis for our classifier - using the standard data split, with 80\% allocated for training and 20\% for testing. We fine-tuned the SciBERT embedding model on the provided data split, where the test part served as a validation subset for the training. The main goal was to fine-tune the embeddings to generalize over unseen sentences similar to the annotated ones. Therefore, the primary evaluation metric at this stage was the validation loss. 
Overall, we achieved a 0.42 validation loss, which indicates that the model should be generalizable to the unlabeled sentences. The accuracy of the final SciBERT model is 0.90. The fine-tuning was performed over three epochs, with the batch size 8 for training and evaluation subsets. Identical data splits with the same sentence samples were applied to train logistic regression, SVM, and XGBoost models. 5-fold cross-validation was performed to identify the most robust and consistent machine learning model. Of the three, logistic regression showed the best performance score, achieving 0.94 accuracy, 0.93 F1, recall, and precision scores derived from the evaluation of the test data.

\subsection*{Classifying the Central Research Questions}
\label{matmethods_rqs}
For our second measure of societal orientation, we sought to determine what the main research question of each paper was, and if it was societally-oriented. We first used a Llama3.1 model (8 billion features) to extract the main research question for each paper, prompted with the paper's title and abstract. We randomly sampled 200 papers for manual validation, to ascertain that the extracted research question was relevant and valid. Then, we utilized our sentence-level classifier to label which research question was societally-engaging. The full prompt is found in Appendix D. 

\subsection*{Topic Modeling}

To identify the main themes expressed in societally-oriented content, we applied topic modeling using the BERTopic package \cite{grootendorst2022bertopic}. The input corpus consisted solely of sentences classified as societally-oriented by our classifier.

Preprocessing involved part-of-speech tagging with the spaCy library \cite{spacy2}, retaining only nouns, verbs, and adverbs to focus on semantically meaningful tokens, while discarding stopwords and other less-informative terms. Additonally, entries with fewer than three words were discarded to enhance quality and computational efficiency.

We then followed the standard BERTopic pipeline. Sentence embeddings were generated using the \textit{all-mpnet-base-v2} SentenceTransformer model \cite{reimers-2019-sentence-bert}. A CountVectorizer limited to unigrams (n-gram range = (1,1)) and configured with a maximum document frequency of 0.8 was used to construct the term-document matrix.

Dimensionality reduction was performed using UMAP (Uniform Manifold Approximation and Projection) with 20 neighbors, five output components, and cosine distance as the similarity metric. Clustering was then conducted via HDBSCAN (Hierarchical Density-Based Spatial Clustering of Applications with Noise), using a minimum cluster size of 300 and default settings for other parameters. HDBSCAN identifies dense regions in the embedding space and assigns low-density points to noise, resulting in coherent and interpretable topic clusters.  Once clusters were established, the most representative terms for each cluster were extracted to characterize the resultant topics.

\newpage
\bibliographystyle{apsr}
\bibliography{references}

% \pagebreak
\clearpage
\appendix

\section*{Appendix A: Robustness Checks}

\begin{figure}[htbp]
\centering
\includegraphics[width=0.8\linewidth]{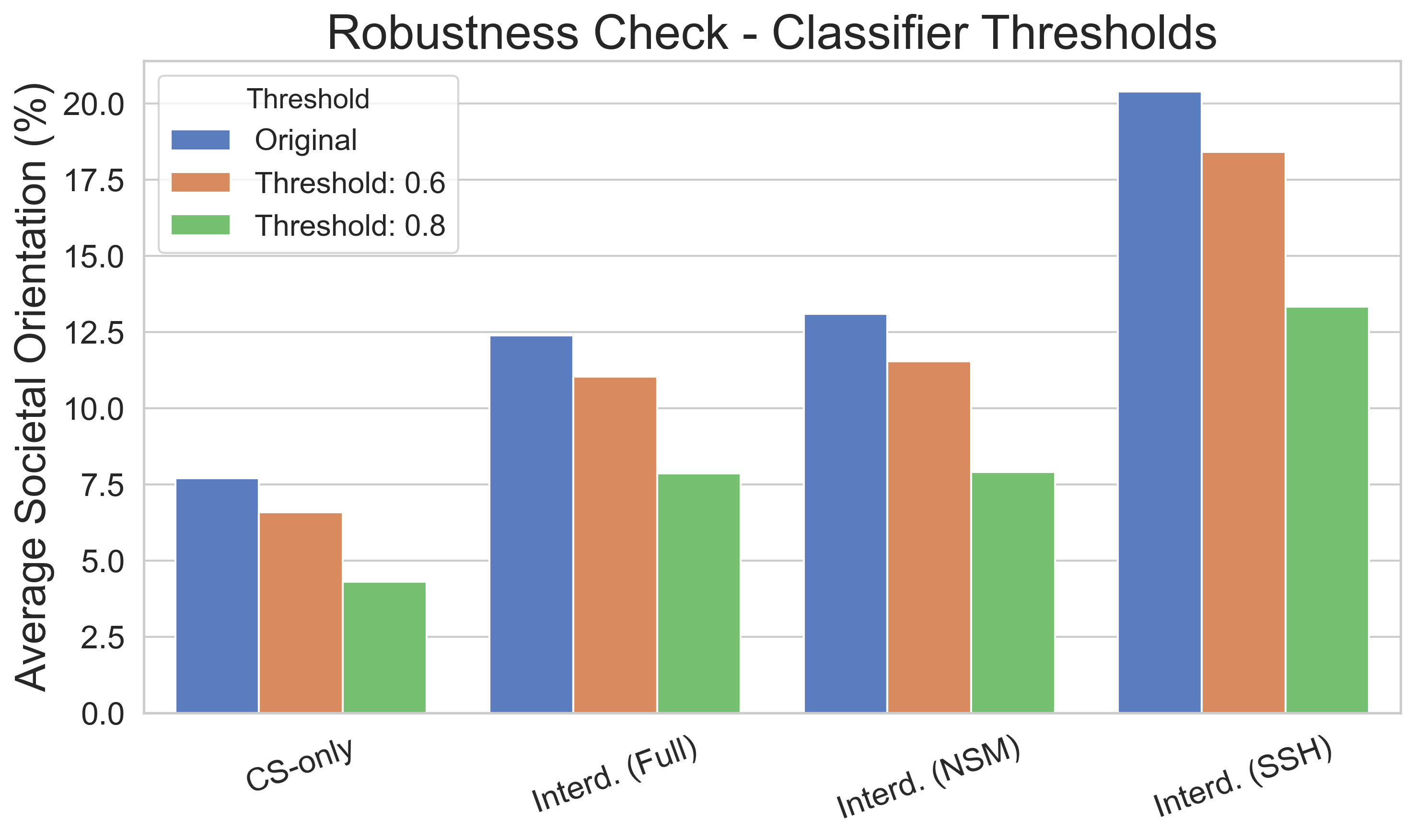}
\caption{\textbf{Classifier Threshold Sensitivity.} Average societal orientation (\%) by team type across three sentence-level classification thresholds: 0.5 (original), 0.6, and 0.8. Results show consistent ordering across team types and highlight the robustness of interdisciplinary effects, particularly for teams involving SSH researchers.}
\label{fig:threshold_robustness}
\end{figure}

\paragraph*{Threshold Sensitivity of Societal Orientation Classifier.}
To assess the robustness of our classifier-based measure of societal orientation, we tested its sensitivity to alternative probability thresholds for sentence classification. In addition to the default threshold of 0.5, we re-estimated our main regression models using stricter thresholds of 0.6 and 0.8. This helps assess whether our findings hold under more conservative assumptions about what constitutes societal orientation. As shown in Figure~\ref{fig:threshold_robustness}, the relative differences between team types remain consistent across thresholds. CS-only teams consistently exhibit the lowest levels of societal orientation, while interdisciplinary teams—particularly those involving social scientists or humanists—continue to demonstrate the highest levels. For example, at the original 0.5 threshold, SSH-inclusive teams average 20.4\% societally oriented content, compared to just 7.7\% for CS-only teams. Even at the strictest threshold of 0.8, this gap remains substantial (13.3\% vs. 4.3\%). These findings reinforce the stability of our classifier and support the robustness of our conclusions across varying levels of conservativeness in sentence classification.

\paragraph*{Field of Study Thresholds.}
\label{fos_robustness}
To assess the robustness of our interdisciplinary classification scheme, we varied the threshold used to assign an author’s primary field of study based on their publication history. Our default analysis used a 90\% threshold, requiring that at least 90\% of an author’s publications fall within a single disciplinary category in order to be labeled as single-discipline. We then compared this to more inclusive thresholds of 80\% and 75\%, which allow more mixed-background authors to be classified as interdisciplinary. As a result, the share of teams labeled as interdisciplinary increases as the threshold is lowered.

\begin{figure}[htbp]
\centering
\includegraphics[width=0.8\linewidth]{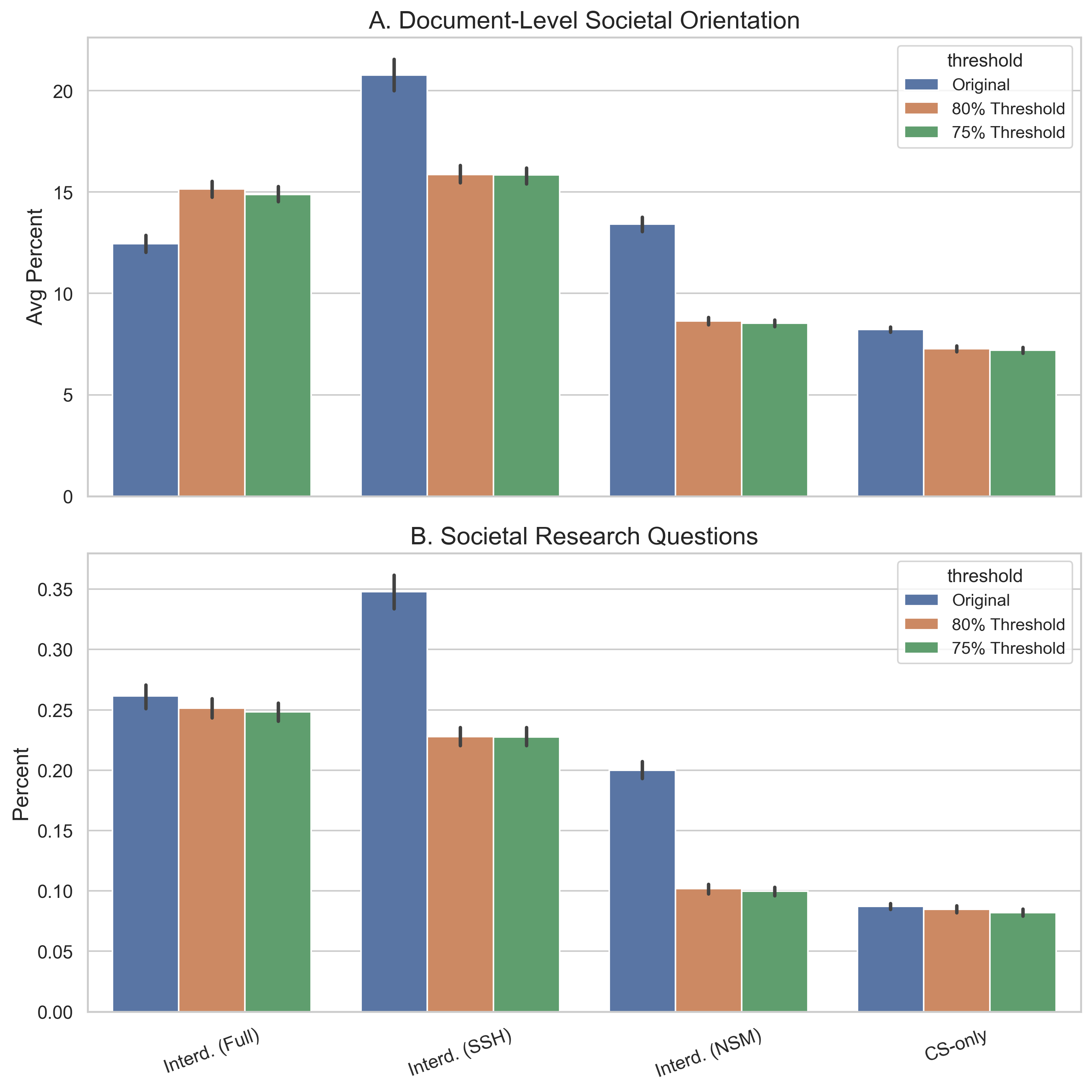}
\caption{
\textbf{Field-of-Study Assignment Thresholds.} 
Average societal orientation (\%) by team type across three thresholds (90\%, 80\%, 75\%) used to classify authors' primary field based on their prior publications. 
Lowering the threshold makes it easier for authors with mixed publication records to be categorized as interdisciplinary, increasing the number of interdisciplinary teams. 
Despite these changes, interdisciplinary teams consistently show higher societal orientation than CS-only teams, underscoring the robustness of the observed effects.}

\label{fig:field_thresholds_robustness}
\end{figure}

As shown in Figure~\ref{fig:field_thresholds_robustness}, our main findings remain robust across all three thresholds. At the document level (Panel A), interdisciplinary teams consistently exhibit higher levels of societal orientation than CS-only teams. For instance, at the original threshold, SSH-inclusive teams average 20.8\% societally oriented content per paper, compared to 8.2\% for CS-only teams. Even under the more inclusive 75\% threshold, this gap persists (15.8\% vs. 7.2\%).

The same trend holds when examining whether the main research question of each paper is societally focused (Panel B). SSH-inclusive teams lead at all thresholds, with 34.8\% of their research questions classified as societally oriented under the default threshold, compared to just 8.7\% for CS-only teams. These findings confirm that our core claims are not sensitive to how authors’ disciplinary identities are defined and remain robust even under more expansive definitions of interdisciplinarity.

\section*{Appendix B: Regression Analysis of Societal Orientation}

\subsection*{Fixed Effects Model Controlling for Subfield and Time}

To verify that the observed relationship between interdisciplinary collaboration and societal orientation is not confounded by structural factors, we estimated a fixed-effects panel regression model. The dependent variable was the percentage of societally-oriented sentences in each paper, and the model included controls for publication year, ArXiv subfield, team size, and article length. Standard errors were clustered to account for heteroskedasticity.

As shown in Table~\ref{tab:fixed_effects}, all interdisciplinary team types were significantly associated with higher societal orientation relative to CS-only teams. Fully interdisciplinary teams showed an increase of 4.99 percentage points in societal content ($p < 0.001$), SSH-inclusive teams showed the largest effect at 5.57 points ($p < 0.001$), and NSM-inclusive teams showed a 3.30-point increase ($p < 0.001$). These results reinforce the conclusion that interdisciplinary team composition—especially when it includes researchers from the social sciences and humanities—is a strong and consistent predictor of societal orientation in AI research.

\begin{table}[h!]
\centering
\caption{Effect of Interdisciplinary Team Composition on Societal Orientation (Fixed Effects PanelOLS)}
\label{tab:fixed_effects}
\begin{tabular}{lrrrr}
\toprule
\textbf{Variable} & \textbf{Coefficient} & \textbf{Robust SE} & \textbf{t-stat} & \textbf{P-value} \\
\midrule
Fully Interdisciplinary & 4.995 & 0.217 & 23.04 & $<$0.001 \\
Interdisciplinary (NSM) & 3.303 & 0.162 & 20.43 & $<$0.001 \\
Interdisciplinary (SSH) & 5.571 & 0.295 & 18.92 & $<$0.001 \\
Team Size & 0.124 & 0.026 & 4.68 & $<$0.001 \\
Article Length (words) & -0.0003 & 0.00002 & -14.63 & $<$0.001 \\
\bottomrule
\end{tabular}
\end{table}

\subsubsection*{Model Statistics}
\begin{itemize}
    \item $R^2$ (within) = 0.368
    \item $R^2$ (between) = 0.139
    \item $R^2$ (overall) = 0.349
    \item Time periods = 11 (years)
\end{itemize}

\subsection*{Robustness to Classifier Thresholds}

To test the robustness of our societal orientation metric, we varied the probability threshold used to classify sentences. In addition to the default threshold of 0.5, we tested stricter thresholds of 0.6 and 0.8. Across all thresholds, interdisciplinary teams remained significantly more likely to produce societally-oriented research. The relative ordering of team types remained stable, with SSH-inclusive teams consistently showing the strongest effect.

\begin{table}[h!]
\centering
\caption{Interdisciplinary Effects across Classifier Thresholds}
\label{tab:classifier_thresholds}
\begin{tabular}{lrrr}
\toprule
\textbf{Variable} & \textbf{Threshold 0.5} & \textbf{Threshold 0.6} & \textbf{Threshold 0.8} \\
\midrule
Fully Interdisciplinary   & 4.92 & 4.51 & 3.37 \\
Interdisciplinary (NSM)   & 3.32 & 2.93 & 1.93 \\
Interdisciplinary (SSH)   & 5.32 & 4.88 & 3.53 \\
Team Size                 & 0.12 & 0.10 & 0.06 \\
Article Length (words)    & -0.0003 & -0.0003 & -0.0002 \\
\bottomrule
\end{tabular}
\begin{flushleft}
\footnotesize \textit{Note}: All coefficients are significant at $p < 0.001$. \\
Table~\ref{tab:classifier_thresholds}. Interdisciplinary Effects across Classifier Thresholds.
\end{flushleft}
\end{table}

\subsection*{Robustness to Field of Study Assignment Thresholds}

We also examined whether our results were sensitive to how authors were assigned a primary field of study. Our default classification used a 90\% threshold, requiring that at least 90\% of an author’s prior publications fall within one field. We tested alternative thresholds of 80\% and 75\%, which label more mixed-background authors as interdisciplinary. As shown in Table~\ref{tab:field_thresholds}, results remain stable across thresholds: interdisciplinary teams consistently exhibit higher societal orientation compared to CS-only teams, with minor variation in magnitude.

\begin{table}[h!]
\centering
\caption{Effect of Interdisciplinary Team Composition under Alternative Field-of-Study Thresholds}
\label{tab:field_thresholds}
\begin{tabular}{lrr}
\toprule
\textbf{Team Type} & \textbf{80\% Threshold} & \textbf{75\% Threshold} \\
\midrule
Fully Interdisciplinary   & 3.23 & 3.04 \\
Interdisciplinary (NSM)   & 1.52 & 1.43 \\
Interdisciplinary (SSH)   & 2.72 & 2.61 \\
\bottomrule
\end{tabular}
\begin{flushleft}
\footnotesize \textit{Note}: PanelOLS coefficient estimates with entity and year fixed effects. The dependent variable is the share of societally-oriented content. The reference category is CS-only teams. \\
Table~\ref{tab:field_thresholds}. Effect of Interdisciplinary Team Composition under Alternative Field-of-Study Thresholds.
\end{flushleft}
\end{table}

\appendix
\section*{Appendix C: Annotation Codebook for Societal Engagement}

\subsection*{Objective}

As public discourse on AI's societal implications grows, this study seeks to determine whether these concerns are receiving meaningful exploration by the researchers driving technological advancements. To explore this, we annotate AI research papers by focusing on the abstract, introduction, and conclusion sections—where authors typically state the motivation behind the research, considerations informing design choices, and the potential future impact of the work. This annotation task involves identifying sentences from computer science research papers that show societal influence, meaning they contain ethical or societal considerations rather than just technical improvements. The focus is on recognizing instances where research aligns with broader societal needs or ethical standards.

\subsection*{Definition of Societal Engagement}

Societal engagement in research refers to the integration of ethical values and societal concerns that guide, shape, and impact the motivations, design, and potential outcomes of a study. We define societal engagement as encompassing two components:

\begin{itemize}[leftmargin=*, topsep=0pt, itemsep=3pt]
    \item \textbf{Ethical values related to AI}: Principles that promote responsible research practices in AI by prioritizing harm prevention, individual rights, and societal welfare. 
    \textit{Examples}: Transparency, Human Rights, Explainability, Fairness, Accountability, Regulation, Safety, Human Oversight, Ethics, Bias, Privacy, Data Protection, Beneficence, Justice, Respect for Law, Respect for Persons, Autonomy.
    
    \item \textbf{Societal issues}: Publicly salient topics that connect research outcomes to pressing public concerns. These often reflect broader societal debates and policy implications. 
    \textit{Examples}: Job automation, Misinformation, Cybersecurity, Surveillance, Autonomous vehicles/weapons, Human meaning, Social interaction, Economic inequality, Climate change, Radicalization, Hate speech, Public health, Pandemics, Government regulation, Policy.
\end{itemize}

A sentence is considered societally engaged if it discusses ethical principles, specific societal concerns, policy recommendations, or broader impacts of the research (e.g., mitigating or responding to issues). \textbf{Note}: Since all sentences originate from computer science research papers, even a brief mention of a societal aspect (e.g., political tweets as data) qualifies as societal engagement.

In Table \ref{tab:examples_societal},  we list several demonstrative examples of Societal vs. Non-Societal Sentences.

\begin{table}[htbp]
\centering
\caption{Examples to Clarify Distinction Between Societal and Non-Societal Sentences}
\label{tab:examples_societal}
\begin{tabular}{|p{7cm}|c|p{5.5cm}|}
\hline
\textbf{Text} & \textbf{Is\_Societal} & \textbf{Explanation} \\
\hline
``This model prioritizes transparency by ensuring interpretability across various demographic subgroups, allowing stakeholders to assess potential biases in decision-making processes.'' & Yes & Highlights ethical values (transparency, bias) important for trust and fairness. \\
\hline
``We discuss the implications of our model for automated content moderation, acknowledging potential privacy concerns due to extensive surveillance on social media platforms.'' & Yes & Connects to societal issues of privacy and surveillance. \\
\hline
``Our model leverages a fine-tuned BERT with contextual embeddings to reduce error rates in coreference resolution tasks.'' & No & Purely technical improvement, no societal reference. \\
\hline
``This tool could support mental health monitoring, assisting in early intervention efforts.'' & Yes & Public health impact noted, clearly societal. \\
\hline
``The proposed architecture achieves state-of-the-art accuracy on the benchmark dataset, outperforming existing models by a margin of 2.5\%.'' & No & Focus on technical performance, no ethical or societal mention. \\
\hline
\end{tabular}
\end{table}

We recognize that these distinctions are sometimes subtle. Consider:

\begin{itemize}[leftmargin=*, topsep=0pt, itemsep=3pt]
    \item \textbf{Non-societal example}:\\
    ``We employ a hybrid transformer-based approach to improve entity recognition in low-resource languages, with a primary emphasis on enhancing linguistic accuracy.''\\
    Although it addresses low-resource settings, it lacks reference to societal motivations or ethical concerns.

    \item \textbf{Societal example}:\\
    ``Our approach aims to mitigate linguistic biases within sentiment analysis systems, particularly those that impact underrepresented dialects, thereby enhancing equitable access to NLP tools.''\\
    Here, societal engagement is expressed through mention of ``bias,'' ``underrepresentation,'' and ``equitable access.''
\end{itemize}

\subsection*{Special Note on Bias}

The concept of ``bias'' requires special care. When a paper mentions algorithmic or selection bias in a purely technical context, it is \textbf{not} considered societal engagement. Bias is only annotated as societal when explicitly linked to social contexts—e.g., gender, race, or other demographic dimensions.

\appendix
\section*{Appendix D: Extracting the Research Questions}

To extract the main research question from each paper, we used the following prompt with the Llama 3.1 (8B) model via the LangChain interface. The model was queried with temperature set to 0 for replicability.

\begin{quote}
Task: You are an AI assistant trained to extract the main research question from an academic research article based on its title and abstract. 

Given the title and abstract below, provide the main research question in a clear, concise, and question-based format.

Title: \{title\} \\
Abstract: \{abstract\}

Output Format: \\
Provide the main research question as a single, well-formed question that captures the key focus of the study.

OUTPUT: \{\{question\}\}
\end{quote}

We applied this prompt to all papers in the dataset. To validate the quality of extracted questions, we manually reviewed a random sample of 200 results. Each extracted question was then passed through our sentence-level classifier to determine its societal orientation.

\end{document}